\pdfoutput=1

\documentclass[11pt]{article}

\usepackage[blocks]{authblk}
\usepackage[final]{acl}

\usepackage{times}
\usepackage{latexsym}

\usepackage[T1]{fontenc}
\usepackage[utf8]{inputenc}

\usepackage{microtype}
\usepackage{inconsolata}

\usepackage{graphicx}
\usepackage{subcaption}
\usepackage{booktabs}
\usepackage{multirow}
\usepackage{adjustbox}
\usepackage{colortbl}
\usepackage{wrapfig}
\usepackage{makecell}

\usepackage{amsmath}
\usepackage{amssymb}
\usepackage{mathtools}
\usepackage{amsthm}

\usepackage{pifont}
\newcommand{\cmark}{\ding{51}}
\newcommand{\xmark}{\ding{55}}

\usepackage{xcolor}
\usepackage[breakable]{tcolorbox}

\newtcolorbox{keydiscoverybox}{
  breakable,
  colback=gray!3, colframe=gray!50,
  boxrule=0.5pt, arc=4pt,
  left=8pt, right=8pt, top=6pt, bottom=6pt,
}
\newcommand{\keydiscovery}[1]{%
  \begin{keydiscoverybox}\noindent\textbf{Key Discovery.} #1\end{keydiscoverybox}}

\usepackage[capitalize,noabbrev]{cleveref}

\theoremstyle{plain}

\theoremstyle{definition}

\theoremstyle{remark}

\usepackage{soul}
\definecolor{metablue}{RGB}{204,229,255}
\newcommand{\systemname}{BALMS}
\newif\ifhideyw
\hideywtrue
\ifhideyw
  \newcommand{\yw}[1]{}
  \newcommand{\lc}[1]{}
  \newcommand{\yz}[1]{}
  \newcommand{\ap}[1]{}
\else
  \newcommand{\yw}[1]{\sethlcolor{pink}\hl{[Yvonne: #1]}}
    \newcommand{\yz}[1]{\sethlcolor{blue!10}\hl{[Evelyn: #1]}}
  \newcommand{\lc}[1]{\sethlcolor{metablue}\hl{[Leo: #1]}}
  \newcommand{\ap}[1]{\sethlcolor{yellow}\hl{[Arvind: #1]}}
\fi

\usepackage[textsize=tiny]{todonotes}

\title{BALMS: \underline{B}enchmarking \underline{A}gentic \underline{L}LMs for \\ Longitudinal \underline{M}ental Health \underline{S}ensing }

\makeatletter
\renewcommand\AB@affilsepx{   \protect\Affilfont}
\makeatother

\author[1]{\textbf{Yu Yvonne Wu}}
\author[1]{\textbf{Arvind Pillai}}
\author[1]{\textbf{Yuliang Chen}}
\author[2]{\textbf{Yuwei Zhang}}
\author[1]{\authorcr \textbf{Sudarshan Regmi}}
\author[1]{\textbf{Tess Z. Griffin}}
\author[1]{\textbf{Michael V. Heinz}}
\author[1]{\authorcr \textbf{Lisa A. Marsch}}
\author[1]{\textbf{Nicholas C. Jacobson}}
\author[1]{\textbf{Andrew Campbell}}
\affil[1]{Dartmouth College}
\affil[2]{University of Cambridge}
\affil[ ]{\authorcr \small{\texttt{yvonne.wu@dartmouth.edu}}}

\begin{document}
\maketitle

\begin{abstract}

Mental health assessment relies on episodic self-report scales, which convert subjective states such as stress into numerical scores but provide only sparse snapshots of wellbeing. Wearable devices offer longitudinal behavioral and physiological signals for continuous, low-burden monitoring. Recent LLM-driven personal-health agents enable natural language queries over wearable signals, but mainly handle short-term, retrieval-based lookups (\textit{e.g.}, highest step count over a week). They do not evaluate whether agents can reason over long-term signals to predict wellbeing scores paired with evidence-grounded rationales.
To address this gap, we introduce \textbf{\systemname{}}, the first systematic benchmark of LLM-based agentic systems for longitudinal mental health sensing. \systemname{} spans 3 real-world longitudinal datasets, 2 task families (closed-form wellbeing-score prediction and rationale generation auto-graded by an LLM-as-Judge), 3 agentic paradigms evaluated across 5 open- and closed-source LLM backbones. We find that 
zero-shot agents rarely outperform a simple mean baseline, except with stronger backbones or compact, semantically meaningful features. Chain-of-thought prompting improves reasoning-oriented backbones, but does not guarantee temporal grounding or numerical correctness.
Together with more analysis on efficiency and temporal scaling, \systemname{} highlights the need for longitudinal mental health agents that selectively retrieve history, ground temporal evidence, and reason over interpretable behavioral features.

\end{abstract}

\section{Introduction}
\label{sec:intro}

Mental wellbeing, encompassing symptoms of depression, anxiety, and stress, is a growing public health concern~\citep{ferrari2022global}. Traditional assessment relies heavily on episodic clinical visits and self-report questionnaires, which provide only sparse snapshots of an individual’s mental state and place substantial burden on both patients and providers~\citep{samhsa2025nsduh}. In contrast, wearable and mobile devices passively and continuously collect multivariate behavioral and physiological signals such as sleep, activity, or heart rate over weeks to years~\citep{globem, nepal2024capturing, gomes2023survey}. These longitudinal streams offer a continuous window into daily wellbeing, capturing temporal dynamics missed by periodic clinical assessment.

\begin{figure}[t]
    \centering
    \includegraphics[width=0.85\columnwidth]{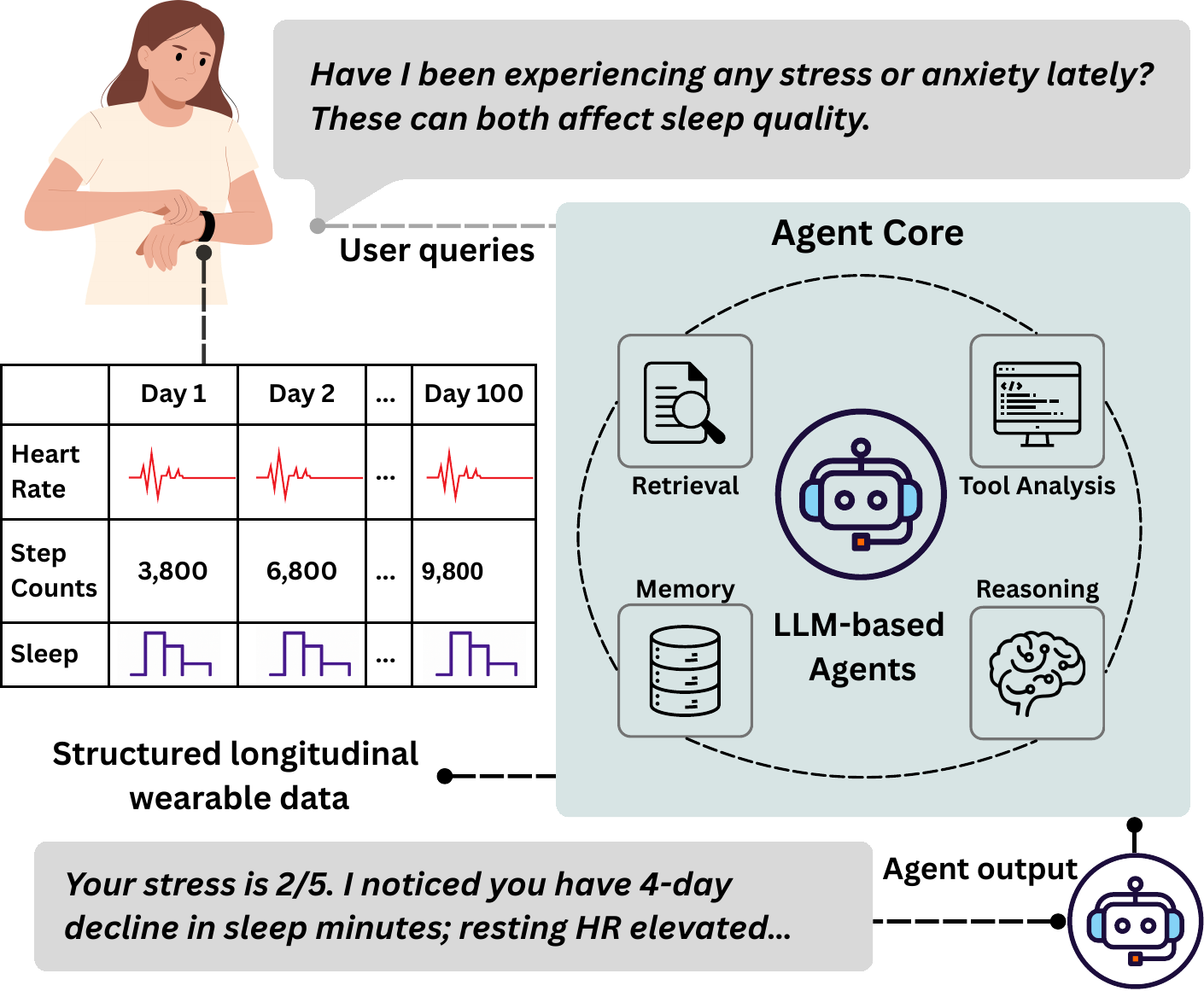}
    \caption{\textbf{LLM-based agents for longitudinal mental health sensing.} Given a multi-channel wearable record and a user query, an agent combines retrieval, memory, tools, and reasoning to output a wellbeing score with an evidence-grounded rationale.} 
    \label{fig:agent_flow}
    \vspace{-15pt}
\end{figure}

Realizing this opportunity requires systems that can transform long, heterogeneous sensing histories into personalized mental health assessments~\citep{ph-llm}.
LLM-based agentic systems offer a promising paradigm for this problem. Standalone LLMs are bounded by fixed context windows and unreliable numerical reasoning over long sensor sequences, so they cannot directly ingest multi-month, multi-channel histories or compute over them. Agentic augmentations such as retrieval, external memory, and tool use address these bottlenecks (Figure~\ref{fig:agent_flow}): they let the model bring in only the relevant past context, delegate numerical computation to executable tools, and produce auditable natural-language rationales rather than a single opaque output~\citep{yao2022react, wang2023augmenting, lewis2021rag}.
Recent agentic systems have shown strong performance in clinical and biomedical reasoning tasks, including diagnostic question answering, evidence retrieval, and multi-turn clinical dialogue~\citep{schmidgall2025agentclinicmultimodalagentbenchmark, arora2025healthbenchevaluatinglargelanguage}, though these benchmarks remain anchored in text-centric modalities. More recently, agentic LLMs have begun to operate directly on wearable and mobile sensing data~\citep{kim2024health, merrill2026transforming, choube2025gloss, heydari2025anatomy}, opening the door to personalized, context-aware outputs grounded in the user's own history~\citep{kim2026codasaicodatascientistbiomarker}. 

However, current wearable-sensing agents remain underexplored along two key axes. First, existing wearable-LLM work targets \emph{short, fixed-window numerical perception} (\textit{i.e.,} factual lookups about a record such as the highest step count in a week or the average resting heart rate over a month~\citep{choube2025gloss, heydari2025anatomy, merrill2026transforming, wu2026wearablefoundationmodelsstatic}).
These tasks test factual lookup and arithmetic over sensor records, but do not evaluate whether an agent can infer mental health states from long-term behavioral patterns. 
Second, other systems explore qualitative case examples or open-ended insights but lack paired wellbeing score prediction explanation, making it difficult to assess whether the generated rationale supports a concrete clinical prediction.

\begin{figure}[t]
    \centering
    \includegraphics[width=\columnwidth]{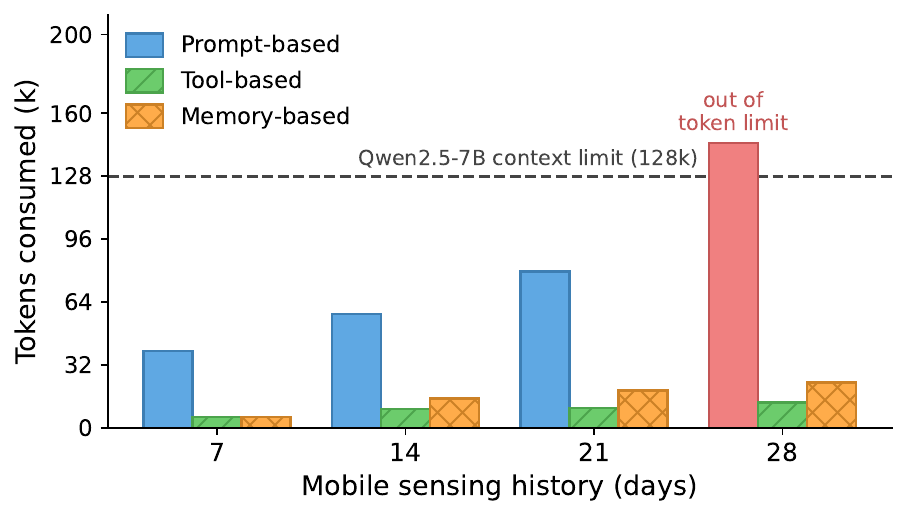}
    \caption{\textbf{Token consumption on stress detection} as wearable sensing duration grows from 7 to 28 days. Prompt-based input exceeds the model's context limit.}
    \label{fig:motivation}
\end{figure}

Therefore, these limitations reflect three technical challenges that arise when scaling agents to longitudinal sensing. 
First, multi-channel sensing histories can quickly exceed standard LLM context windows when high-frequency or long-duration sensor streams are converted into text (Figure~\ref{fig:motivation}); Second, mental health prediction requires numerical grounding over temporal patterns, yet LLMs are known to be unreliable when reasoning directly over long numerical sequences~\citep{pillai2025time2lang, spathis2024first}. Third, wearable and mobile datasets vary substantially in sensor types, feature formats, and observation horizon, meaning that a carefully crafted agentic design that works for one dataset may fail to transfer to another~\citep{luo2025foundationmodelmultivariatewearable}. Together, these challenges make it unclear which agentic paradigm is best suited for long-horizon mental health sensing.

To address this gap, we introduce \systemname{}, a comprehensive benchmark designed to evaluate LLM-based agentic systems for longitudinal mental health sensing. Utilizing three real-world smartphone and wearable datasets (\textit{e.g.}, Fitbit), we systematically investigate agent design space across multiple backbones. Specifically, our contributions are: 
\begin{itemize}
    
    \item \textbf{Task Formulation:} We formalize longitudinal mental health sensing as an agentic benchmark, requiring models to jointly generate verifiable numerical wellbeing scores and evidence-grounded rationales over long-term passive sensing histories. 
    \item \textbf{Unified Evaluation:} We implement and evaluate three core agentic paradigms (prompt-, tool-, and memory-based) across five open- and closed-source LLM backbones, establishing a standardized infrastructure for clinical agent assessment.
    \item \textbf{Empirical Agent Design Insights:} We demonstrate that zero-shot agentic prediction remains challenging, except with stronger backbones or compact, semantically meaningful sensor features, and that chain-of-thought prompting significantly boosts performance on reasoning-tuned backbones. Crucially, our analyses reveal that agents benefit far more from selective memory and semantically meaningful features than from raw sensor streams or extended context windows.

\end{itemize}

\section{Benchmarking LLM Agents for Longitudinal Mental Health Sensing}
\label{sec:method}
\begin{figure*}[!t]
    \centering
    \includegraphics[width=\linewidth]{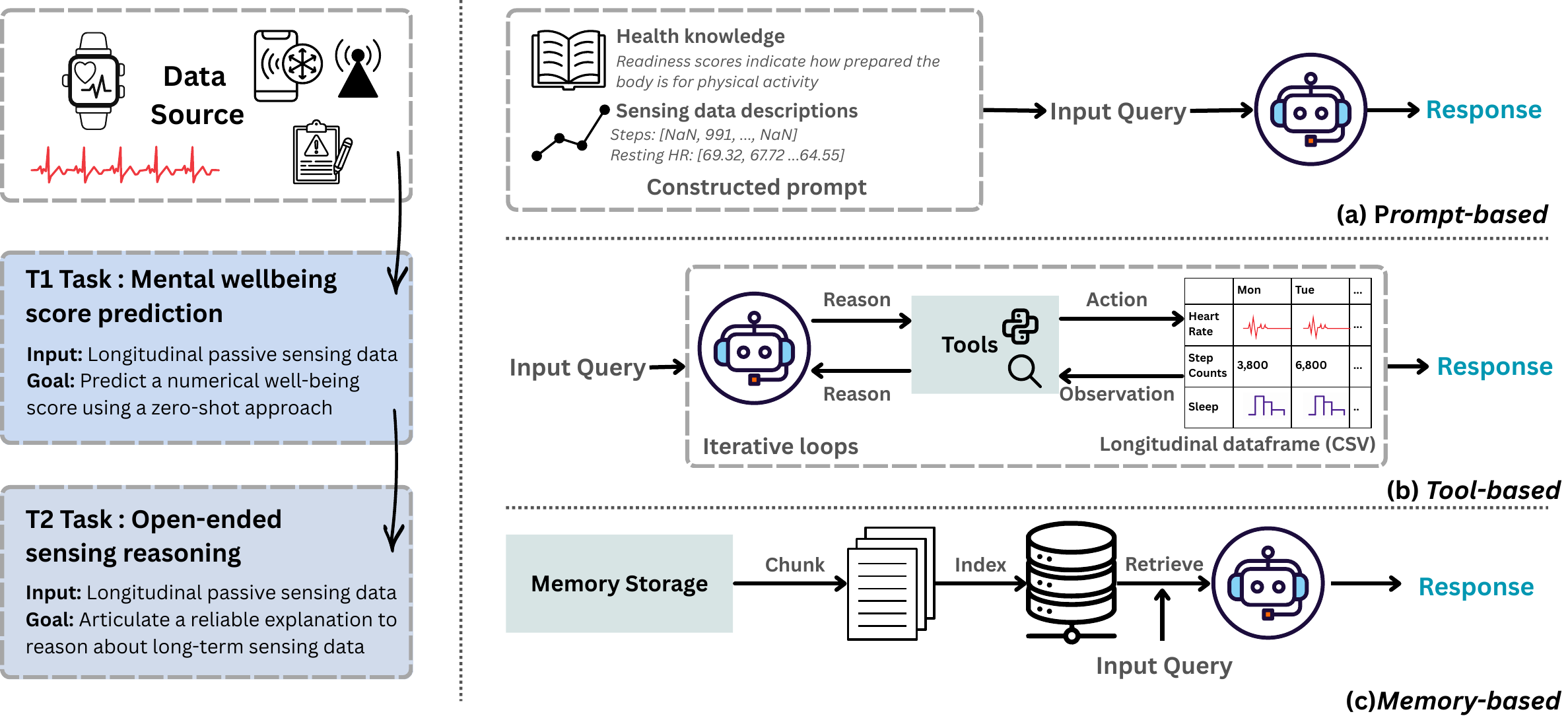}
    \caption{\textbf{Overview of BALMS}, a benchmark for LLM-based agentic systems on longitudinal mental health sensing. Left: two task families share longitudinal passive sensing data as input. Right: the three agentic paradigms applied to both tasks. BALMS evaluates these paradigms across three real-world longitudinal datasets and five LLM backbones, scoring T1 by MAE and T2 by an LLM-as-a-Judge rubric.
    }
    \label{fig:agent-archs}
\end{figure*}
\noindent\textbf{Overview.} To study how different agentic designs handle longitudinal wearable and mobile sensing data, we introduce \systemname{}, a multi-domain benchmark that evaluates agentic LLMs on tasks requiring temporal understanding, wellbeing-score prediction, and evidence-grounded reasoning. \systemname{} is built upon 3 real-world longitudinal passive sensing datasets spanning months to years. Using these datasets, we benchmark two tasks: a closed-ended mental wellbeing score prediction and an open-ended sensing data reasoning. Across both tasks, we evaluate 3 representative agentic paradigms, which differ in how longitudinal data is exposed to the LLM.
\systemname{} is intended for continuous, low-burden monitoring systems that return feedback to users \emph{between} clinical visits rather than for diagnosis, so that users can make sense of the wellbeing score and the textual feedback they receive. Accordingly, strong benchmark performance indicates that an agent's scores agree with self-reports and that its rationales are verifiable against the sensing window.

\subsection{Datasets}
\label{sec:datasets}
We construct \systemname{} using 3 different longitudinal datasets for mental wellbeing monitoring, summarized in Table~\ref{tab:datasets}. 
Detailed data processing is described in Appendix~\ref{app:data}.
\begin{table}[t]
\centering
\setlength{\tabcolsep}{3pt}
\renewcommand{\arraystretch}{1.05}
\begin{adjustbox}{width=\columnwidth}
\begin{tabular}{llll}
\toprule[1.5pt]
Datasets & DiversityOne & PMData& GLOBEM \\
\midrule
Avg Length & 28 days & 5 month & multi years \\
Sensors & 10 & 5 & 8 \\
Participants & 782 & 16 & 497 \\
Sensors & \makecell[l]{Accelerometer,\\ Gyroscope, \\ Wi-Fi, etc} & \makecell[l]{Steps,\\ Resting HR, \\ Burned Calories, etc} & \makecell[l]{Step,\\ Sleep, \\Accelerometer, etc} \\
Tasks & Emotion prediction & Stress detection & Anxiety detection \\
\bottomrule[1.5pt]
\end{tabular}
\end{adjustbox}
\caption{\small{\textbf{Datasets statistics}}}
\label{tab:datasets}
\vspace{-10pt}
\end{table}

\begin{itemize}
    \item \textbf{DiversityOne}~\cite{Busso_2025} is a multi-country smartphone sensing dataset collected from 782 university students across 8 countries over 28 days. It captures phone-derived raw signals (\textit{e.g.}, accelerometer, Wi-Fi traces) alongside a daily evening mood self-report.

    \item \textbf{PMData}~\cite{pmdata} tracks 16 participants over 5 months, pairing Fitbit-derived signals (steps, resting heart rate, calories) with daily PMSys wellness self-reports (stress, fatigue, etc.) scored on a 1--5 Likert scales. We use the daily stress score as our prediction target.

    \item \textbf{GLOBEM}~\cite{globem} is a multi-year longitudinal benchmark from 497 college students that pairs smartphone passive sensing (location, screen status, etc.) and Fitbit-derived activity and sleep tracking with weekly EMA PHQ-4 self-reports, delivered across 10-week study windows for four consecutive years; we use the PHQ-4 anxiety subscale as the prediction target.
\end{itemize}


\subsection{Task Families}
\label{sec:tasks}
Existing agentic systems for wearable analysis evaluate primarily on numerical perception over fixed windows, answering factual lookups about a record such as the highest step count in a week or the average resting heart rate over a month~\citep{merrill2026transforming,choube2025gloss,heydari2025anatomy}. \systemname{} instead targets 
numerical mental-wellbeing scores
over long horizons and organizes the evaluation around two task families: closed-form wellbeing-score prediction (T1) and open-ended rationale generation (T2).

\noindent\textbf{T1: Mental-wellbeing score prediction.}
We formalize this setting as a closed-form regression task: given a target day and a look-back window of multivariate sensing signals (\textit{e.g.}, step count, heart rate, calories), the system outputs a single integer self-report target on the dataset's native Likert scale. Concretely, we predict daily mood on DiversityOne, daily stress on PMData, and the PHQ-4 anxiety subscale on GLOBEM (Section~\ref{sec:datasets}). The formulation follows Health-LLM's zero-shot setting~\citep{kim2024health} and isolates the agent's predictive capability from its capacity to generate text.

\noindent\textbf{T2: Open-ended sensing reasoning.}
A reliable agentic system should not only emit a wellbeing score but also articulate how it arrives at that prediction from longitudinal numerical signals. This second task therefore evaluates an agent's ability to interpret multi-day sensor trajectories and translate them into a justified wellbeing assessment. Going beyond prior work that reports only case-study explanations or open-ended wearable insights without paired score targets~\citep{kim2024health,merrill2026transforming}, we require the agent to produce a free-form chain-of-thought rationale alongside the score, citing the specific sensor channels and values it relied on and reasoning about multi-day patterns such as trends, weekly cycles, or recovery. For example, on PMData's stress task, a competent system might highlight a four-day decline in sleep minutes and an elevated resting heart rate before concluding with a stress score of 4. Because such open-ended outputs cannot be graded by numeric error metrics alone, we score them with an LLM-as-a-Judge rubric, detailed in Section~\ref{sec:eval} and templated in Appendix~\ref{app:prompt:judge}.

\subsection{Benchmarking LLM Systems}
\label{sec:systems}
Figure~\ref{fig:agent-archs} outlines the architectures, and Table~\ref{tab:methods_compare} compares prior work on longitudinal health prediction capabilities. Below, we instantiate one representative method per paradigm.

\begin{table*}[t]
\centering
\setlength{\tabcolsep}{6pt}
\renewcommand{\arraystretch}{1.1}
{\footnotesize
\begin{tabular}{llccccc}
  \toprule[1.5pt]
  \textbf{Paradigm} & \textbf{System}
  & \textbf{Rationale} & \makecell{\textbf{Raw passive}\\\textbf{sensing data}} & \textbf{Tools} & \textbf{Memory} & \makecell{\textbf{Dataset-}\\\textbf{centric}} \\
  \midrule
  \multirow{2}{*}{Prompt-based}
    & Health-LLM~\citep{kim2024health}                       & \xmark & \xmark & \xmark & \xmark & \xmark \\
    & Englhardt et al.~\citep{englhardt2024classification}   & \cmark & \xmark & \xmark & \xmark & \cmark \\
  \midrule
  \multirow{2}{*}{Memory-based}
    & Chunk RAG~\citep{lewis2021rag}                          & \xmark & \xmark & \xmark & \cmark & \xmark \\
    & RAPTOR~\citep{sarthi2024raptor}                        & \xmark & \xmark & \xmark & \cmark & \xmark \\
  \midrule
  \multirow{4}{*}{Tool-based}
    & PHIA~\citep{merrill2026transforming}                   & \cmark & \xmark & \cmark & \xmark & \xmark \\
    & GLOSS~\citep{choube2025gloss}                          & \cmark & \cmark & \cmark & \xmark & \cmark \\
    & PHA~\citep{heydari2025anatomy}                         & \cmark & \xmark & \cmark & \xmark & \cmark \\
    & LifeAgentBench~\citep{tian2026lifeagentbench}          & \cmark & \xmark & \cmark & \xmark & \cmark \\
  \bottomrule[1.5pt]
\end{tabular}
}
\caption{
\textbf{Comparison of representative LLM-based agentic systems for passive mobile sensing.}
\textit{Rationale} = produces a free-form rationale alongside or instead of a closed-form label.
}
\vspace{-10pt}
\label{tab:methods_compare}
\end{table*}

\subsubsection{Prompt-Based Reasoning Systems}
Existing LLM approaches for mental health prediction rely on prompt-based reasoning, directly probing the model's pretrained knowledge. Here, the numerical data is transformed into a text prompt, and the model is asked to produce a prediction~\citep{kim2024health}. These approaches demonstrate that LLMs can perform inference on sensing data to produce predictions and explanations through carefully designed prompts and chain-of-thought reasoning~\citep{gruver2023large}.

\noindent\textbf{Health-LLM}~\citep{kim2024health} is the most comprehensive baseline in this family and is originally evaluated across multiple wearable datasets and a range of mental-wellbeing prediction targets. It formats each sensor modality as a per-day numeric array spanning the look-back window, includes the user's demographic profile and a task-specific instruction (\textit{e.g.}, the score range and the question being asked) as additional context, and assembles them into a single prompt that queries the LLM to output the target label as an integer. We adopt this prompt template directly and only vary the look-back window and the underlying backbone; the full template is provided in Appendix~\ref{app:prompt:healthllm}. Such prompt-only systems consume a fixed context window and do not retrieve memory, call tools, or plan analyses, which motivates the stronger agentic paradigms.

\subsubsection{Tool-Based Agents}
Tool-based methods extend agents with executable tools like code interpreters~\cite{allan2010survey} to extract rolling statistics, trends, and summaries. This allows LLMs to compute over long, multi-channel sensing records rather than reading them directly. While recent tools (\textit{e.g.}, PHIA~\cite{merrill2026transforming}, GLOSS~\cite{choube2025gloss}, PHA~\cite{heydari2025anatomy}) follow this template, they are typically tied to specific sensor schemas and lack evaluation across diverse cohorts.

\noindent\textbf{PHIA}~\cite{merrill2026transforming} is a representative tool-based system that generalizes across passive-sensing setups. It preloads each user's record as a set of in-memory pandas DataFrames and drives the LLM agents through a ReAct loop~\cite{yao2022react}. 
At each step, the model generates a reasoning trace and selects an action from a predefined tool space.
Following PHIA's implementation, this action is instantiated as executable Python code over the preloaded DataFrames, including operations such as date filtering and groupby aggregation.
The execution result is appended to the trajectory as context and fed back into the next step until the agents output a final answer (Appendix~\ref{app:prompt:healthllm}).

\subsubsection{Memory- and Retrieval-Based Agents}
A complementary way to handle long histories is to keep the past in an external memory store and retrieve only the relevant entries into the prompt at inference time~\cite{wang2023augmenting}. The agent then reasons only over the entries of the user's history most relevant to the current query, selected by content similarity rather than recency. To our knowledge, no memory-augmented agent~\citep{chhikara2025mem0buildingproductionreadyai} has been developed for mental health sensing; we therefore start from a standard retrieval-augmented generation (RAG) recipe and adapt it to the wearable-sensing setting. 

\noindent\textbf{Chunk-based RAG.} Standard text RAG chunks documents at the token or paragraph level~\cite{jin2025long}, which does not align with the granularity of mobile sensing, where both the self-reported labels and the sensor aggregates are naturally daily. We therefore chunk memory at the day level rather than at the text-token level. Concretely, each historical day's sensing data are encoded with a sentence-transformer~\cite{reimers2019sentence} into a memory bank. At inference, we retrieve the top-$k$ similar days by cosine similarity and pass only those days, together with the query to the LLM.

\noindent\textbf{Tree-based RAG.}
Chunk-based RAG retrieves only flat day-level entries, which can miss intra-day structure as well as longer-horizon context. We therefore adapt RAPTOR~\citep{sarthi2024raptor} to the daily sensing setting with a two-level memory tree. Leaves are fine-grained sub-day records that capture intra-day variation in the sensing stream, for example half-hour slots on PMData and four time-of-day segments on GLOBEM. Each daily internal node is an LLM-generated abstractive summary over its day's leaves. At inference, retrieval operates on the collapsed tree, where every node (leaf or daily summary) is a candidate, so the top-$k$ entries returned to the LLM can mix fine-grained sub-day evidence with daily abstractions. Tree-construction and retrieval settings for both memory-based systems, including the per-dataset sub-day granularity and the top-$k$ used at inference, are reported in Appendix~\ref{app:data}.

\section{Experimental Setup}

\subsection{LLM Backbones}
We evaluate each system across five LLM backbones: three open-source instruct models (\emph{Qwen2.5-7B/14B-Instruct}~\citep{Yangetal2025} and \emph{Mistral-7B-Instruct-v0.3}~\citep{jiang2023mistral7b}), one open-source reasoning-distilled model (\emph{DeepSeek-R1-Distill-Qwen-14B}~\citep{deepseekai2025r1}), and one closed-source model (\emph{Claude-Haiku-4.5}~\citep{haiku2025}).
\begin{table*}[!t]
\centering
\setlength{\tabcolsep}{6pt}
\renewcommand{\arraystretch}{1.05}
{\footnotesize
\begin{tabular}{llcccccc}
  \toprule[1.5pt]
  \multirow{2.5}{*}{\textbf{Backbone}} & \multirow{2.5}{*}{\textbf{Method}}
  & \multicolumn{2}{c}{\textbf{DiversityOne(Mood)}} & \multicolumn{2}{c}{\textbf{PMData(Stress)}} & \multicolumn{2}{c}{\textbf{GLOBEM(Anxiety)}} \\
  \cmidrule(lr){3-4}\cmidrule(lr){5-6}\cmidrule(lr){7-8}
  & & MAE$^\downarrow$ & +CoT MAE$^\downarrow$
  & MAE$^\downarrow$ & +CoT MAE$^\downarrow$
  & MAE$^\downarrow$ & +CoT MAE$^\downarrow$ \\
  \midrule\midrule
  
  \textcolor{gray}{Reference baseline} & \textcolor{gray}{Mean Predictor }
    & \multicolumn{2}{c}{\textcolor{gray}{0.58}}
    & \multicolumn{2}{c}{\textcolor{gray}{0.48}}
    & \multicolumn{2}{c}{\textcolor{gray}{0.80}} \\
  \midrule\midrule
  \multicolumn{8}{l}{\small \textit{Open-source LLMs}} \\
  \midrule
  \multirow{4}{*}{Qwen2.5-7B-Instruct}
    & Health-LLM & 0.60 & 0.62 & 0.65 & 0.55 & 1.17 & 1.30 \\
    & RAG        & 0.96 & 0.99 & 0.72 & 0.63 & 1.20 & 1.37 \\
    & Raptor     & 1.15 & 1.29 & 0.78 & 0.64 & 1.21 & 1.22 \\
    & PHIA       & 1.22 & 1.46 & 0.64 & 0.48 & 1.48 & 1.54 \\
  \midrule
  \multirow{4}{*}{Mistral-7B-Instruct-v0.3}
    & Health-LLM & 1.06 & 1.58 & 0.88 & 1.07 & 1.83 & 2.00 \\
    & RAG        & 1.52 & 1.67 & 0.61 & 2.08 & 1.28 & 2.58 \\
    & Raptor     & 1.61 & 2.15 & 0.83 & 2.12 & 1.21 & 1.96 \\
    & PHIA       & 0.84 & 0.91   & 0.63 & 1.57 & 1.29 & 2.40 \\
  \midrule
  \multirow{4}{*}{Qwen2.5-14B-Instruct}
    & Health-LLM & \underline{0.44} & 0.79 & 0.62 & 0.52 & \underline{0.86} & 1.02 \\
    & RAG        & 0.53 & 0.84 & 0.71 & 0.47 & 0.90 & 1.19 \\
    & Raptor     & 1.16 & 1.68 & 0.76 & 0.48 & 0.93 & 1.05 \\
    & PHIA       & 2.11 & 3.15   & 0.57 & 0.52 & 1.52 & 1.88 \\
  \midrule
  \multirow{4}{*}{\makecell{DeepSeek-R1-Distill-\\Qwen-14B}}
    & Health-LLM & 0.49 & \underline{0.46} & 0.65 & 0.55 & 1.23 & 1.17 \\
    & RAG        & 0.48 & \textbf{0.41} & 0.58 & 0.53 & 1.11 & 1.05 \\
    & Raptor     & 1.10 & 1.07 & 0.59 & 0.50 & 1.21 & 1.21 \\
    & PHIA       & 0.99 & 0.97 & \textbf{0.50} & \underline{0.46} & 1.68 & 1.29 \\
  \midrule\midrule
  \multicolumn{8}{l}{\small \textit{Closed LLMs}} \\
  \midrule
  \multirow{4}{*}{Claude-Haiku-4.5}
    & Health-LLM & \textbf{0.42} & \textbf{0.41} & 0.64 & 0.48 & \textbf{0.83} & \textbf{0.81} \\
    & RAG        & 0.87 & 0.51 & 0.67 & 0.56 & 0.96 & 0.93 \\
    & Raptor     & 0.98 & 0.73 & 0.74 & 0.59 & 0.89 & \underline{0.86} \\
    & PHIA       & 1.11 & 0.82 & \underline{0.54} & \textbf{0.45} & 1.41 & 1.27 \\
  \bottomrule[1.5pt]
\end{tabular}
}
\caption{
\textbf{Zero-shot mental health detection.} We compare 4 agentic systems with 5 backbones reporting MAE without and with CoT. No paradigm wins across all three datasets.  Best per column in \textbf{bold}, second-best \underline{underlined}.
}

\label{tab:main:retrieval}
\end{table*}

\subsection{Benchmark Evaluation Metrics}
\label{sec:eval}
For the closed-form prediction task defined in Section~\ref{sec:tasks}, we report mean absolute error (MAE) between the predicted integer self-report and the gold-standard label on each dataset's native Likert scale, averaged over all held-out target days.

\noindent\textbf{T2: LLM-as-Judge for open-ended reasoning.}
We evaluate whether the generated rationale is both temporally meaningful and grounded in the sensing evidence. We use a single Llama-3.3-70B-Instruct judge, which is given the sensing window, gold self-report score, agent prediction, and generated rationale.
This allows the judge to assess not only whether the explanation is fluent, but also whether its claims are supported by the input window and consistent with the predicted score.


The rubric evaluates two tiers. Tier 1 (Temporal Reasoning) adapts TemporalBench~\citep{cai2024temporalbench} to evaluate six distinct operations over longitudinal records: C1 Alignment (mapping temporal conditions to data fields), C2 Slicing (comparing temporal segments), C3 Difference Judgment (quantifying relative change), C4 Lag (delayed effects), C5 Structure (identifying peaks, troughs, or change points), and C6 Interaction (reasoning about joint cross-channel effects). Dimensions are scored as correct, incorrect, or not invoked, the latter prevents penalizing rationales for omitting unnecessary operations. We report invocation rate, conditional accuracy given invocation, and coverage (the rate of invoked-and-correct dimensions). Tier 2 (General Quality) adapts the PHIA expert rubric~\citep{merrill2026transforming} to score five dimensions on a 1--5 Likert scale: Faithfulness (whether the prediction follows logically from the reasoning), Evidence Grounding (verifiability of cited facts against the input window), Domain Knowledge, Safety, Clarity, and an Overall holistic score. The full prompt is provided in Appendix~\ref{app:prompt:judge}.

\subsection{Implementation Details}
\label{sec:setup}
For all systems, we evaluate the same target-day splits and use identical input windows for fair comparison. Open-source backbones are served locally with vLLM on 4 A6000 GPUs. Claude-Haiku-4.5 is queried through the Anthropic API using the same prompt formats and decoding settings. We follow official implementations of Health-LLM and PHIA and set the maximum ReAct iterations to 10. Chunk RAG follows official LangChain implementations and RAPTOR builds a two-level memory tree over sub-day records and daily summaries. 



\section{Results and Discussion}
\label{sec:results}

\begin{figure*}[!t]
    \centering
    \includegraphics[width=0.95\linewidth]{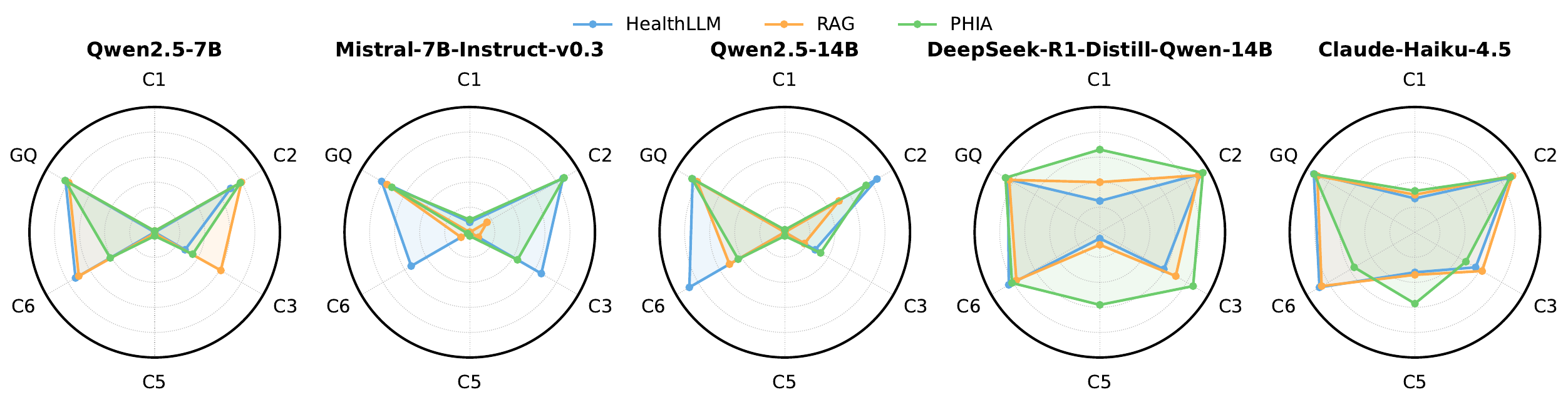}
    \caption{\textbf{LLM-as-Judge results on GLOBEM.} Each subplot corresponds to one backbone, overlaying 3 agentic systems; 
    \emph{GQ} is the Tier-2 general-quality overall Likert score, rescaled to $[0,1]$ for comparability. Higher is better on all axes. Only Claude-Haiku-4.5 closes both the rationale-quality and temporal-grounding gaps, while open-source instruct backbones produce fluent rationales but struggle with temporal reasoning and rarely invoke alignment (C1) or structure (C5).}
    \label{fig:judge_radar_globem}
    \vspace{-10pt}
\end{figure*}

\subsection{Zero-Shot Mental-Wellbeing Prediction}
\label{sec:zero-shot}
\keydiscovery{
Existing zero-shot agents rarely outperform a simple mean baseline unless the backbone is stronger or the input features are compact and semantically meaningful.
}

\noindent This task evaluates whether agentic systems can predict daily mental health labels from longitudinal, multimodal sensing histories without task-specific training. As a reference baseline, we use a mean predictor that outputs the average training set label without utilizing sensing inputs.

Table~\ref{tab:main:retrieval} shows that zero-shot agentic prediction is feasible, but its effectiveness varies by LLM backbone and agentic paradigm.
Larger, closed-source, or reasoning-tuned backbones are more likely to approach or surpass the mean predictor.
For example, Claude-Haiku-4.5 with Health-LLM achieves 0.42 MAE on DiversityOne, outperforming the 0.58 mean baseline, while PHIA reaches 0.54 MAE on PMData, close to the 0.48 baseline.
However, smaller models like Mistral-7B frequently fail to match the mean predictor. These results suggest that existing agents can extract useful behavioral signals from longitudinal records, but much of the predictive power in daily self-reports is still captured by simple dataset-level priors.
      

Scaling the backbone mainly benefits prompt-based and memory-based agents: both Health-LLM and RAG systems improve performance with larger backbones, since they rely on interpreting textualized sensing histories. Conversely, PHIA does not scale consistently. This indicates that tool-based agents are not limited solely by language capacity, but by whether generated code can operate over the target schema. Specifically, PHIA excels only on PMData, where Fitbit-style aggregates match its design. This approach fails on DiversityOne and GLOBEM, where raw mobile streams make generated code brittle. Thus, tool-mediated reasoning requires the tool interface to match the underlying sensing data structure. Appendix~\ref{app:phia-failures} provides a detailed failure analysis.

\noindent\textbf{Design insight.} Future longitudinal sensing agents should pair capable backbones with either structured sensor representations or raw sensing signals rather than relying on zero-shot prompting alone.



\subsection{Reasoning and Rationale Analysis}
\keydiscovery{
Chain-of-thought helps reasoning-oriented backbones, but larger backbones do not guarantee temporal grounding or numerical correctness.}
\noindent Beyond base T1 score prediction, we examine whether explicit reasoning improves both predicted scores and T2 rationale quality. Table~\ref{tab:main:retrieval} shows that CoT prompting primarily benefits reasoning-tuned backbones. With DeepSeek and Claude architectures, +CoT consistently reduces MAE across most configurations, yielding improvements up to 41.4\%. This suggests that reasoning-focused models better organize temporal evidence before scoring. Figure~\ref{fig:judge_radar_globem} shows that temporal-reasoning capability is gated by the backbone, and only the frontier-lab model closes both the quality and the grounding gaps. Claude-Haiku-4.5 reaches the highest overall rationale quality while consistently invoking every temporal reasoning dimension.

While agents using instruct backbones frequently produce fluent explanations, they rarely invoke temporally grounded evidence. Specifically, they fail to execute precise operations over time series, such as accurately slicing historical windows or tracking cross-channel correlations between distinct sensing streams (i.e., C1 alignment and C5 structure invocation are near zero). This suggests that many open-source agents imitate the surface form of reasoning without reliably grounding it in the underlying time series. 

\noindent\textbf{Reasoning case study.}
To analyze detailed reasoning traces against each rubric dimension, we walk through a single GLOBEM window predicted by two backbones and annotate the rationales against the Tier-1 and Tier-2 axes in Appendix~\ref{app:case-study}.

\noindent\textbf{Design insight.} Future mental health agents should explicitly build in temporal operations for comparing behavioral baselines, detecting sustained changes, and analyzing sensor interactions.

\subsection{Long-Horizon Scaling}
\label{sec:long-horizon-scaling}

\begin{figure}[!t]
    \centering
    \includegraphics[width=\columnwidth]{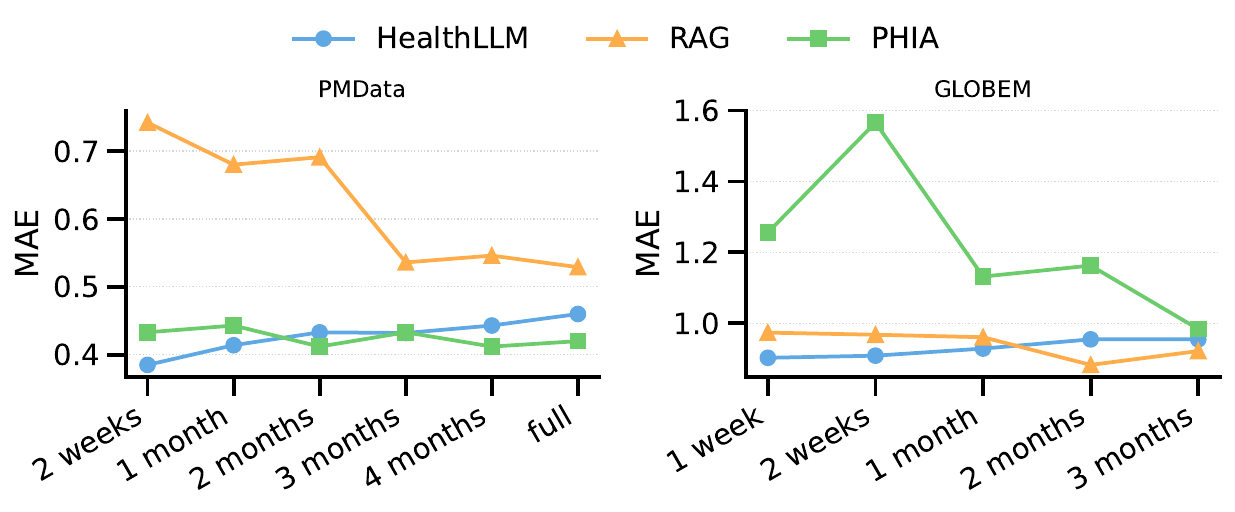}
    \caption{\textbf{Effect of look-back window length} on MAE for Qwen2.5-14B across PMData and GLOBEM. Corresponding latency curves are reported in Figure~\ref{fig:latency_lookback}. Health-LLM degrades as the window grows, RAG benefits the most, and PHIA stays stable but remains schema-sensitive on GLOBEM.}
    \label{fig:lookback}
    \vspace{-10pt}
\end{figure}

This section evaluates whether agentic systems remain effective across different temporal scales of sensing context. We vary the look-back window on PMData and GLOBEM using Qwen2.5-14B-Instruct while keeping the target day fixed. 

Figure~\ref{fig:lookback} shows that temporal-scale robustness differs substantially across paradigms. Health-LLM degrades as the look-back window increases, suggesting that simply adding more textualized records can dilute target-day evidence and increase numerical reasoning difficulty~\cite{spathis2024first}. RAG benefits more consistently from longer histories, especially on PMData, where MAE decreases $\sim$29\% as more historical records become available for retrieval. The gain is smaller on GLOBEM ($\sim$9\%), likely because its sparser passive-sensing features provide less informative retrieved evidence than PMData’s Fitbit-style aggregates. PHIA remains stable on PMData because preloaded DataFrames decouple its tool execution from prompt length. However, it remains weaker and noisier on GLOBEM, again reflecting the schema sensitivity observed in earlier sections. Its robustness also comes with higher latency, since multi-step code execution remains costly across all horizons as shown in Appendix Figure~\ref{fig:latency_lookback}.


\noindent\textbf{Design insight.} Future clinical agents must extend retrieval beyond context reduction, leveraging it for temporal reasoning over personal baselines and behavioral shifts.


\subsection{Efficiency and Sensitivity Analysis}
\begin{figure}[!t]
    \centering
    \includegraphics[width=\columnwidth]{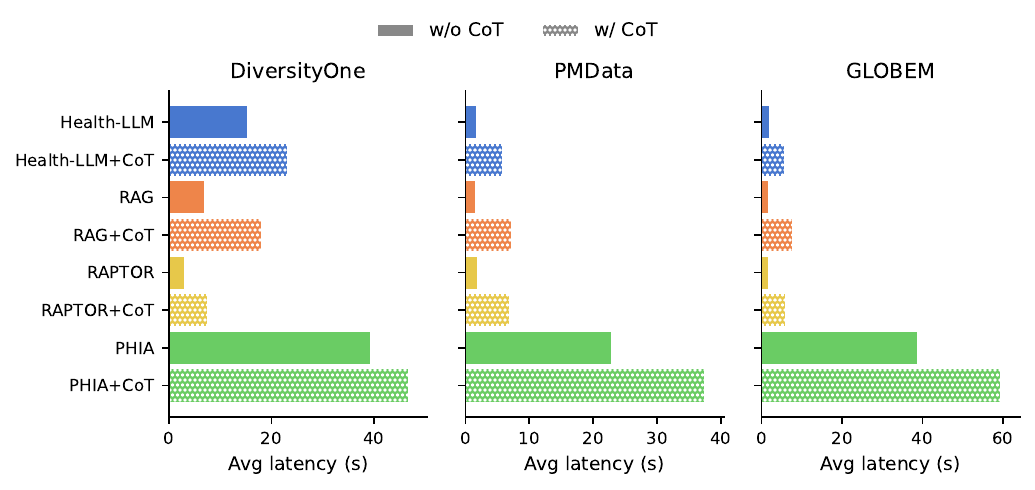}
    \caption{\textbf{Average per-sample latency} of each agentic system with Qwen2.5-14B-Instruct. 
    }
    \label{fig:latency_panel_qwen14b}
    \vspace{-10pt}
\end{figure}

As shown in Figure~\ref{fig:latency_panel_qwen14b}, prompt- and memory-based paradigms (Health-LLM and RAG) are highly efficient as they bypass recursive execution workflows. Conversely, tool-based agents (PHIA) introduce an extensive latency penalty, particularly when paired with CoT prompting, due to the multi-step overhead of generating, running, and debugging code iteratively. However, expanding raw input signals does not uniformly improve predictive power. Our sensor sensitivity analysis on GLOBEM (Table~\ref{tab:sensor_sensitivity}) demonstrates that dropping mobile streams to evaluate a compact, smartwatch-only configuration (Fitbit) frequently matches or surpasses the full multimodal baseline. This highlights a key limitation: contemporary LLMs excel at reasoning over high-level, semantically rich behavioral primitives but struggle to extract signal directly from raw, high-frequency data streams. More analyses on other datasets are in Appendix~\ref{app:latency}.


\begin{table}[!t]
\centering
\setlength{\tabcolsep}{2.8pt}
\renewcommand{\arraystretch}{1.02}
{\scriptsize
\begin{tabular}{llcccc}
  \toprule[1.2pt]
  \multirow{2.5}{*}{\textbf{Backbone}} & \multirow{2.5}{*}{\textbf{Method}}
  & \multicolumn{2}{c}{\textbf{Full}} & \multicolumn{2}{c}{\textbf{Fitbit}} \\
  \cmidrule(lr){3-4}\cmidrule(lr){5-6}
  & & MAE$^\downarrow$ & +CoT$^\downarrow$
  & MAE$^\downarrow$ & +CoT$^\downarrow$ \\
  \midrule\midrule
  \multirow{4}{*}{\makecell[l]{Qwen2.5-\\7B-Instruct}}
    & Health-LLM & 1.17 & 1.30 & 1.12 & 1.24 \\
    & RAG        & 1.20 & 1.37 & 1.43 & 1.27 \\
    & Raptor     & 1.21 & 1.22 & 1.22 & 1.20 \\
    & PHIA       & 1.48 & 1.54 & 1.18   & 1.08   \\
  \midrule
  \multirow{4}{*}{\makecell[l]{Mistral-7B-\\Instruct-v0.3}}
    & Health-LLM & 1.83 & 2.00 & 1.31 & 1.85 \\
    & RAG        & 1.28 & 2.58 & 1.36 & 2.11 \\
    & Raptor     & 1.21 & 1.96 & 1.21 & 1.96 \\
    & PHIA       & 1.29 & 2.40 & 1.18   & 1.59   \\
  \midrule
  \multirow{4}{*}{\makecell[l]{Qwen2.5-\\14B-Instruct}}
    & Health-LLM & 0.86 & 1.02 & 0.80 & 1.14 \\
    & RAG        & 0.90 & 1.19 & 0.90 & 1.01 \\
    & Raptor     & 0.93 & 1.05 & 0.93 & 1.05 \\
    & PHIA       & 1.52 & 1.88 & 1.05   & 1.17   \\
  \midrule
  \multirow{4}{*}{\makecell[l]{DeepSeek-R1-\\Distill-Qwen-14B}}
    & Health-LLM & 1.23 & 1.17 & 1.15 & 1.17 \\
    & RAG        & 1.11 & 1.05 & 1.10 & 1.06 \\
    & Raptor     & 1.21 & 1.21 & 1.21 & 1.13 \\
    & PHIA       & 1.14   & 1.07   &  1.08   & 0.97   \\
  \bottomrule[1.2pt]
\end{tabular}
}
\caption{\textbf{Sensor-set sensitivity on GLOBEM.} Each agentic system on GLOBEM under two sensor configurations: \textit{Full} (mobile + smartwatch) vs.\ \textit{Fitbit} (smartwatch only), w/o and w/cot. Dropping raw mobile streams and keeping only Fitbit semantic rich features often matches or beats the full setup.}
\label{tab:sensor_sensitivity}
\vspace{-15pt}
\end{table}



\section{Related Work}
\label{sec:related_work}
A growing body of benchmarks has begun to evaluate LLM-based agents in healthcare and mobile environments~\citep{schmidgall2025agentclinicmultimodalagentbenchmark, arora2025healthbenchevaluatinglargelanguage, tian2026lifeagentbench}. For example, MedAgentBench evaluates medical agents in an interactive FHIR-based electronic health record environment~\citep{jiang2025medagentbench}. Mobile-agent benchmarks evaluate whether agents can control smartphone applications through graphical interfaces~\citep{deng2024mobilebench}. These benchmarks are valuable for measuring tool use, planning, and interaction. However, they do not capture the core challenges of longitudinal mental-health sensing: reasoning over multi-week to multi-year wearable histories, grounding predictions in numerical temporal patterns, and evaluating whether generated explanations support concrete wellbeing outcomes.


\section{Conclusion}
We introduced \systemname{}, a benchmark for evaluating LLM-based agentic systems on longitudinal mental-health sensing. Across three passive-sensing datasets, two task families, and three agentic paradigms, BALMS shows that zero-shot agents can achieve competitive wellbeing-score prediction without task-specific training, but reliable performance depends on how agents access and ground longitudinal evidence. Our analyses further show that chain-of-thought prompting benefits reasoning-oriented backbones, while longer histories and additional sensors help only when agents can selectively retrieve relevant context and interpret semantically meaningful features. These findings highlight the need for agent designs with selective memory, numerical grounding, and schema-aware reasoning for longitudinal mental-health sensing.

\section*{Limitations}
\systemname{} focuses on representative agentic paradigms rather than exhaustively covering all possible agent designs. In particular, we evaluate single-agent prompt-based, tool-based, and memory-based systems, but do not study multi-agent collaboration, planner-executor architectures, or agents that dynamically combine tools, memory, and reflection. Our evaluation also focuses on retrospective prediction from existing datasets, and does not test agents in real-world interactive or intervention settings. For rationale evaluation, we use an LLM-as-Judge rubric to provide scalable assessment of temporal grounding, numerical correctness, and explanation quality. While this enables systematic comparison across many systems, it does not replace expert clinical or human-subject evaluation. Future work should incorporate clinician and user assessments, evaluate agents in prospective deployment, and study safety, calibration, and personalization under real-world mental-health support scenarios.

\section*{Ethics Statement}
\textbf{Potential Misuse.}
While \systemname{} benchmarks agentic frameworks to advance continuous health monitoring, we state that these systems are intended as clinical decision-support tools and not as autonomous diagnostic replacements for professional medical care. Misuse of such models through autonomous, uncalibrated mental health profiling could lead to severe outcomes, including false reassurance or unwarranted clinical anxiety.

\noindent\textbf{Data Privacy.} This work relies entirely on the secondary analysis of previously collected, fully de-identified, and publicly or restricted-access research cohorts (DiversityOne, PMData, and GLOBEM). Because the data are completely anonymized and publicly accessible for research purposes, this study does not constitute human subjects research under standard institutional guidelines, requiring no institutional review board (IRB) review. No protected health identifiers (PHIs) were accessed, stored, or processed at any stage of our benchmarking pipeline.

\noindent\textbf{Population Representation.} We acknowledge that passive sensing patterns vary substantially across different demographic, cultural, and socioeconomic cohorts. Models evaluated on these benchmarks risk inheriting or amplifying data-sparsity patterns or behavioral biases inherent to specific student or regional populations. We strongly caution against generalizing these specific zero-shot weights to broader, heterogeneous clinical settings without proactive calibration and local alignment.


\section*{Acknowledgments}
The authors acknowledge support for this research from Evergreen: A Generative AI \& Behavioral Sensing Digital Ecosystem to Promote Student Wellness and Flourishing. This work is made possible through philanthropic gifts to Dartmouth College dedicated to advancing AI-supported well-being and flourishing of college students.


\bibliography{main}

\clearpage
\appendix
\section{Data Processing Details}
\label{app:data}

For the closed-form wellbeing-score prediction task (T1), we follow the zero-shot setup of Health-LLM~\citep{kim2024health} and use a dataset-specific look-back window of daily-aggregated multivariate sensing signals, except where the per-dataset analysis in Section~\ref{sec:long-horizon-scaling} explicitly varies the window length. The look-back is 7 days on DiversityOne (whose total observation horizon is only 28 days) and 14 days on PMData and GLOBEM.

\noindent\textbf{DiversityOne.} \footnote{\url{https://datascientia.disi.unitn.it/projects/diversityone/}} The raw release contains 782 participants spanning eight countries with substantially uneven mood self-report response rates. To control for self-report sparsity, we restrict all evaluations to the Mongolia cohort, which has the highest daily mood-answering rate among the eight countries, yielding 1{,}355 (target-day, user) samples.

\noindent\textbf{PMData.}\footnote{\url{https://datasets.simula.no/pmdata/}} We use all 16 participants and the daily PMSys stress self-report (1--5 Likert) as the prediction target, yielding 1{,}448 samples.

\noindent\textbf{GLOBEM.}\footnote{\url{https://physionet.org/content/globem/1.1/}} Following Health-LLM, we restrict evaluation to the INS-W\_1 cohort and use the PHQ-4 anxiety subscale (0--3) as the prediction target, yielding 1{,}640 samples. We note one scope limitation: although the full GLOBEM release spans four consecutive years, data are collected as separate 10-week study terms; we therefore evaluate only one continuous term and only the anxiety subscale. Across-term discontinuities (changing participants, calibration drifts, and term-specific behavioral baselines) would introduce additional noise and modeling challenges, which we leave to future work.

\begin{table*}[!t]
\centering
\setlength{\tabcolsep}{4pt}
\renewcommand{\arraystretch}{1.05}
{\footnotesize
\begin{tabular}{llcc}
  \toprule[1.2pt]
  \textbf{Dataset} & \textbf{Task (Likert)} & \textbf{Look-back} & \textbf{\# Samples} \\
  \midrule
  DiversityOne & Mood (1--5)            &  7 days & 1{,}355 \\
  PMData       & Stress (1--5)          & 14 days & 1{,}448 \\
  GLOBEM       & PHQ-4 Anxiety (0--3)   & 14 days & 1{,}640 \\
  \bottomrule[1.2pt]
\end{tabular}
}
\caption{\small{\textbf{Processed evaluation set per dataset.} For each dataset we list the prediction target, the default look-back window used for T1 main-table runs, and the number of (target-day, user) samples after cohort selection and missing-label filtering.}}
\label{tab:data_processing}
\end{table*}

\paragraph{Implementation details.} All baselines are run following official guidelines. For the prompt-based system, we adopt the Health-LLM template verbatim. For the tool-based system (PHIA), the ReAct loop is capped at 10 iterations per query, and the in-memory pandas dataframe is initialized with the user's full longitudinal record. For the memory-based systems, Chunk RAG encodes each historical day with a sentence-transformer and retrieves the top-$3$ most similar days using the standard LangChain framework, concatenating them with the query day before passing the prompt to the inference LLM. RAPTOR builds a two-level memory tree over the same records: leaves are sub-day segments (half-hour slots on PMData, four time-of-day segments on GLOBEM), and each daily internal node is an LLM-generated abstractive summary of that day's leaves. Retrieval operates on the collapsed tree with the same top-$3$ budget, so leaves and daily summaries compete as candidates. Open-source models are served via vLLM on 4$\times$ A6000 GPUs.

\paragraph{Tier-2 (general-quality) judge scores.}
Per-cell breakdown of the Tier-2 Likert dimensions on GLOBEM is reported in Table~\ref{tab:judge_tier2}.
\begin{table*}[t]
\centering
\setlength{\tabcolsep}{6pt}
\renewcommand{\arraystretch}{1.05}
{\footnotesize
\begin{tabular}{llccccc}
  \toprule[1.5pt]
  \textbf{Backbone} & \textbf{System}
  & \textbf{Faithfulness} & \textbf{Evidence} & \textbf{Domain} & \textbf{Clarity} & \textbf{Overall} \\
  \midrule\midrule
  \multirow{3}{*}{Qwen2.5-7B}
    & Health-LLM & 3.98 & 4.55 & 4.29 & \underline{4.79} & 4.11 \\
    & RAG        & 3.92 & 3.24 & 4.25 & 4.55 & 3.96 \\
    & PHIA     & 4.07 & 3.50 & 4.43 & 4.71 & 4.13 \\
  \midrule
  \multirow{3}{*}{Mistral-7B-Instruct-v0.3}
    & Health-LLM & 3.08 & 4.61 & \underline{4.45} & 4.65 & 4.06 \\
    & RAG        & 1.46 & 4.03 & 4.43 & 4.70 & 3.81 \\
    & PHIA     & 2.98 & 3.05 & 4.05 & 4.30 & 3.59 \\
  \midrule
  \multirow{3}{*}{Qwen2.5-14B}
    & Health-LLM & 4.14 & \textbf{4.77} & 4.39 & \textbf{4.86} & 4.25 \\
    & RAG        & 4.03 & 4.00 & 4.25 & 4.68 & 4.04 \\
    & PHIA     & \textbf{4.28} & 3.63 & \textbf{4.55} & 4.74 & \underline{4.28} \\
  \midrule
  \multirow{3}{*}{DeepSeek-R1-Distill-Qwen-14B}
    & Health-LLM & 4.11 & 4.58 & 4.30 & 4.36 & 4.23 \\
    & RAG        & 4.13 & 4.60 & 4.23 & 4.30 & 4.18 \\
    & PHIA     & \underline{4.18} & \underline{4.75} & 4.41 & 4.42 & \textbf{4.35} \\
  \bottomrule[1.5pt]
\end{tabular}
}
\caption{\small{
\textbf{Tier-2 (general-quality) Likert scores on GLOBEM, 1--5 scale.}
Each row is one (backbone, agentic system) cell. \textit{Safety} is omitted as every cell scores $5.0$.
Best per column in \textbf{bold}; second-best \underline{underlined}.
}}
\label{tab:judge_tier2}
\end{table*}

\section{PHIA Failure Analysis}
\label{app:phia-failures}

Section~\ref{sec:zero-shot} reported that PHIA does not benefit from
stronger backbones and is competitive only on PMData. This appendix examines
the underlying execution traces to identify why. All numbers below use
Qwen2.5-14B-Instruct, the strongest open-source backbone in our suite, on
seed~0. Findings are consistent across the other backbones (Mistral-7B,
Qwen2.5-7B, DeepSeek-R1-Distill-Qwen-14B).

\subsection{Predictions Collapse to a Single Label}
\label{app:phia-mode-collapse}

Across all three datasets, PHIA emits the same label for a large majority of
target days, regardless of the user's sensor history
(Table~\ref{tab:phia-mode-collapse}). On PMData the modal label (\texttt{3})
is close to the dataset's mean stress, so MAE remains competitive
(0.57 vs.\ mean predictor 0.48); on GLOBEM the modal label (\texttt{2}) is far
from the low-skewed PHQ-4 anxiety distribution and MAE doubles
(1.52 vs.\ 0.80); on DiversityOne the agent fails to parse a valid answer
in almost every window. The apparent success on PMData therefore reflects
agreement between a mode-collapsed prior and the dataset mean, not
genuine grounding in the wearable record.

\begin{table*}[!t]
\centering
\small
\begin{tabular}{lrrr}
\toprule
Dataset & Modal label & Mode rate & MAE \\
\midrule
GLOBEM (anxiety)        & 2     & 85.9\% (1409/1640) & 1.52 \\
PMData (stress)         & 3     & 77.4\% (1132/1463) & 0.57 \\
DiversityOne (mood)     & None  & 99.5\% (757/761)   & 2.11 \\
\bottomrule
\end{tabular}
\caption{PHIA mode-collapse rates with Qwen2.5-14B-Instruct (seed~0). The
modal label is the single most-frequent prediction across all held-out
windows. ``None'' on DiversityOne indicates a non-parseable final answer.}
\label{tab:phia-mode-collapse}
\end{table*}

\subsection{Failure Taxonomy}
\label{app:phia-taxonomy}

Inspecting the traced subset of windows (82 traced on GLOBEM, 10 traced on
PMData), we observe five recurring failure modes that together explain the
mode-collapse behavior above. Counts below are reported on Qwen2.5-14B-Instruct.

\begin{description}
\item[F1.\ Silent code execution.] The model emits a tool\_code block that
either omits a \verb|print()| call or is truncated mid-expression. The
sandbox executes it and returns ``Code executed (no output)''. The agent
then proceeds as if the computation had succeeded, citing numerical claims
that were never produced. On GLOBEM, traced windows averaged 2.0 empty
observations per window (168 events across 82 windows); on PMData, 0.7
per window.

\item[F2.\ Lost state across \texttt{[Act]} blocks.] The model assumes
variables defined in a previous step remain in scope. When they do not,
subsequent blocks raise \texttt{NameError}. We observed
\texttt{avg\_steps}, \texttt{filtered\_df}, \texttt{last\_14\_days},
\texttt{total\_steps}, \texttt{anxiety\_score}, and \texttt{score}
referenced before redefinition. The agent typically responds by retrying
with an even longer code block, which often re-triggers~F1.

\item[F3.\ Force-finish with hallucinated rationale.] When the step budget
expires without a parseable answer, the harness force-finishes by asking
the model to emit a final label. The model writes a fluent rationale
referencing specific deltas, thresholds, or trends that were never actually
computed during the trace. 8 out of 10 traced PMData windows ended this way.

\item[F4.\ Magic-number thresholds.] The model invents thresholds without
deriving them from the user's prior distribution (e.g.\
\texttt{steps > 20000}, \texttt{sleep\_minutes > 1680} on GLOBEM), then
collapses to one of two precomputed branches. This is the proximal cause
of mode collapse: irrespective of the underlying signal, almost every user
falls on the same side of the hard-coded threshold.

\item[F5.\ Schema-blind aggregation on raw streams.] On DiversityOne, the
sub-day DataFrame contains $\sim$18 thirty-minute slots per day. The agent
applies \texttt{.mean()} or \texttt{.sum()} over slot-level rows without
first re-indexing to daily granularity, conflating slot averages with
daily summaries. The same code template that works on PMData's pre-aggregated
Fitbit daily rows produces uninterpretable values on raw mobile streams.
\end{description}

These failure modes are not independent. F1 and F2 produce traces with no
verifiable numerical evidence; the agent then either invokes F4 (a
prior-driven heuristic) or runs out of budget and triggers~F3. F5 explains
why DiversityOne is the worst case: the schema gap means even successful
code execution produces semantically wrong aggregates.

\subsection{Illustrative Traces}
\label{app:phia-traces}

We include two abbreviated traces drawn directly from
\texttt{react\_traces\_pmdata/} and \texttt{react\_traces\_globem/}.
Both use Qwen2.5-14B-Instruct, seed~0.

\paragraph{Trace 1 — PMData (apparent success via force-finish).}
Participant \texttt{p01}, target day 2019-11-25.

\begin{quote}\small
\textbf{[Step 1, Act]} Filters \texttt{daily\_df} to the last seven days,
computes \texttt{mean\_steps}, \texttt{mean\_activity\_minutes},
\texttt{mean\_sleep\_efficiency}; identifies anomalous days. \emph{Code
block is cut off mid-conditional; no \texttt{print()}.} \\
\textbf{[Step 1, Observe]} Code executed (no output). \\
\textbf{[Step 2, Act]} ``Let me refine the approach\ldots''. Re-attempts
the same anomaly check; again truncated mid-expression. \\
\textbf{[Step 2, Observe]} Code executed (no output). \\
\textbf{[Step 3, Act]} Third attempt; same truncation pattern. \\
\textbf{[Step 3, Observe]} Code executed (no output). \\
\textbf{[Step 4, Act]} Final attempt; \texttt{predicted\_stress} branch
truncated before assignment. \\
\textbf{[Step 4, Observe]} Code executed (no output). \\
\textbf{[FORCE-FINISH]} Reasoning: \emph{``The target day showed a
significant decrease in steps and total activity minutes compared to the
previous week's average\ldots sleep efficiency was notably lower than
usual\ldots self-reported mood was also elevated.''} \textbf{Answer: 4}
\end{quote}

\noindent None of the deltas cited in the rationale were ever computed; the
agent's predicted label is essentially a prior-driven guess, formatted as
post-hoc justification. This trace counts as ``successful'' in our aggregate
MAE because a parseable integer was returned (F1+F3).

\paragraph{Trace 2 — GLOBEM (visible failure via state loss).}
Participant \texttt{INS-W\_208}, target day 2018-05-16.

\begin{quote}\small
\textbf{[Step 1, Act]} Emits two consecutive code blocks; the second
references \texttt{total\_steps} defined only in the first. The agent also
hardcodes thresholds: \texttt{step\_threshold = 20000},
\texttt{sleep\_minute\_threshold = 1680}. \\
\textbf{[Step 2, Observe]} \texttt{Error executing code: name
'total\_steps' is not defined}. \\
\textbf{[Step 3, Act]} Recomputes \texttt{total\_steps} and
\texttt{total\_sleep\_minutes}, but now from \texttt{filtered\_df} — which
was also defined only in step~1. \\
\textbf{[Step 3, Observe]} \texttt{Error executing code: name 'filtered\_df'
is not defined}. \\
\textbf{[Step 4, Act]} Rebuilds \texttt{filtered\_df}, recomputes
aggregates, applies the same hardcoded threshold; predicts \texttt{3}. \\
\textbf{[Step 4, Observe]} \texttt{3}. \\
\textbf{[FORCE-FINISH]} \textbf{Answer: 3}
\end{quote}

\noindent Although this trace eventually executes, the prediction is
determined entirely by the F4 magic-number threshold; the recovered
aggregates serve only to choose between two hardcoded branches. This
pattern, repeated across the held-out set, is the mechanical source of the
86\% mode collapse on GLOBEM.

\subsection{Implications}

The failure modes above are not specific to Qwen2.5-14B. We observe the
same patterns across Mistral-7B and DeepSeek-R1-Distill-Qwen-14B, which is
consistent with the observation in
Section~\ref{sec:zero-shot} that stronger backbones do not improve PHIA's
MAE. The bottleneck is the contract between the model and the execution
sandbox — print discipline, state persistence, and threshold derivation —
rather than language reasoning capacity. This motivates the memory-based
paradigm benchmarked in Section~\ref{sec:long-horizon-scaling}, which
exposes the relevant history to the LLM directly and avoids the
silent-execution surface entirely.

\section{Latency Analysis}
\label{app:latency}

\begin{figure}[!t]
    \centering
    \begin{subfigure}[t]{0.48\columnwidth}
        \centering
        \includegraphics[width=\linewidth]{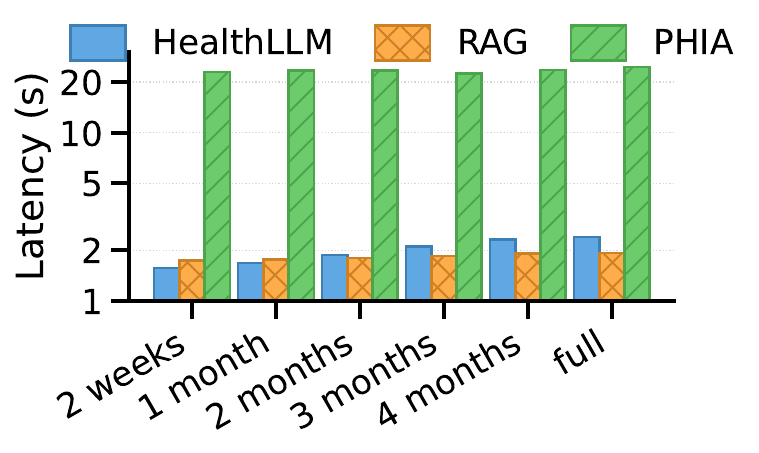}
        \caption{PMData}
        \label{fig:latency_lookback:pmdata}
    \end{subfigure}\hfill
    \begin{subfigure}[t]{0.48\columnwidth}
        \centering
        \includegraphics[width=\linewidth]{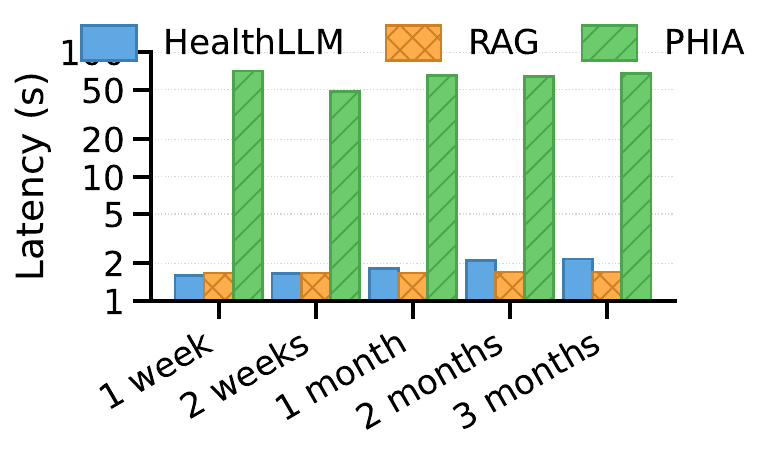}
        \caption{GLOBEM}
        \label{fig:latency_lookback:globem}
    \end{subfigure}
    \caption{Average per-sample latency as a function of look-back window length for Qwen2.5-14B on PMData and GLOBEM.}
    \label{fig:latency_lookback}
\end{figure}

\paragraph{Latency vs.\ look-back window.}
Figure~\ref{fig:latency_lookback} accompanies the look-back analysis in Section~\ref{sec:results} (Figure~\ref{fig:lookback}), reporting average per-sample latency for Qwen2.5-14B-Instruct as the look-back window grows on PMData and GLOBEM.

\begin{figure}[!t]
    \centering
    \includegraphics[width=0.95\linewidth]{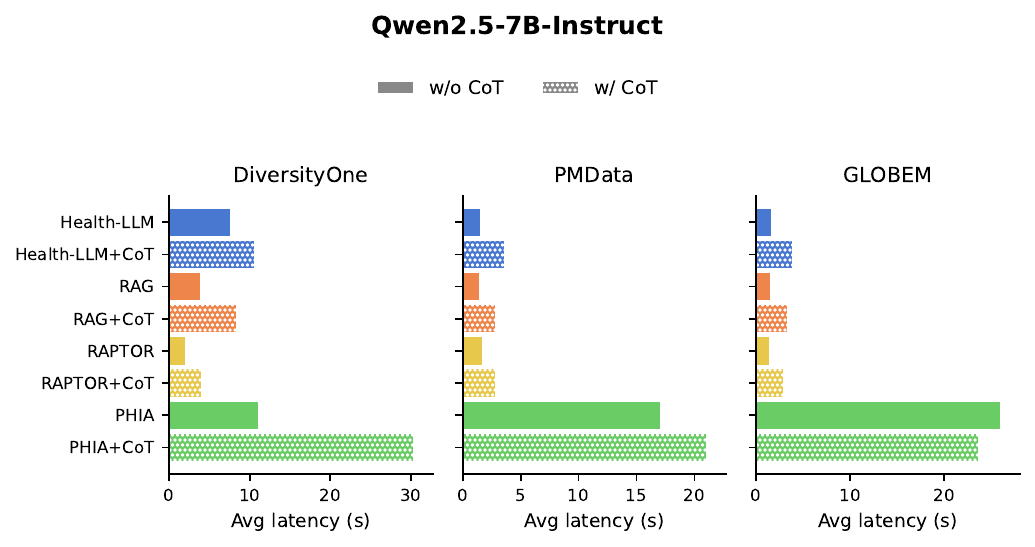}
    \caption{Average per-sample latency of each agentic system with Qwen2.5-7B-Instruct.}
    \label{fig:latency_panel_qwen7b}
\end{figure}

\begin{figure}[!t]
    \centering
    \includegraphics[width=0.95\linewidth]{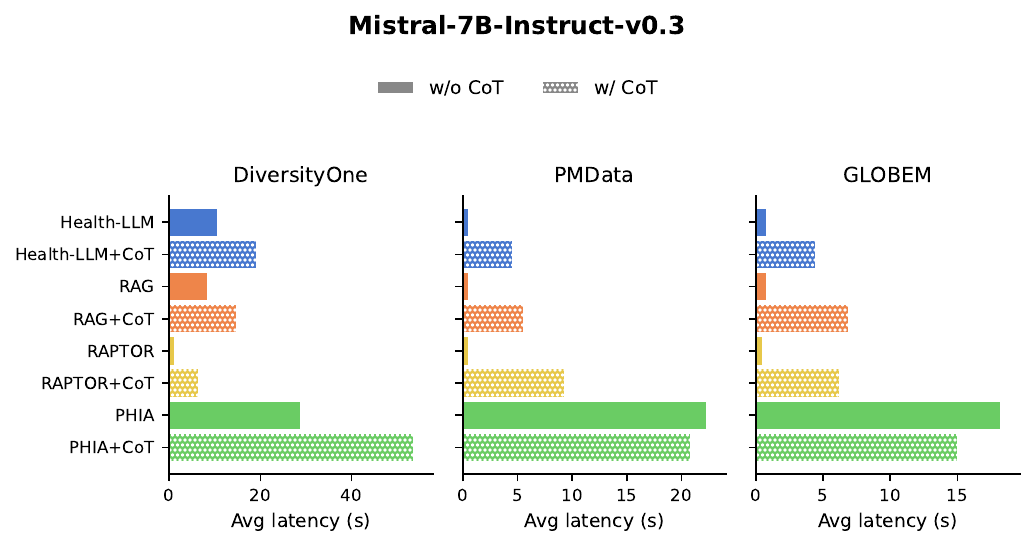}
    \caption{Average per-sample latency of each agentic system with Mistral-7B-Instruct-v0.3.}
    \label{fig:latency_panel_mistral7b}
\end{figure}

\begin{figure}[!t]
    \centering
    \includegraphics[width=0.95\linewidth]{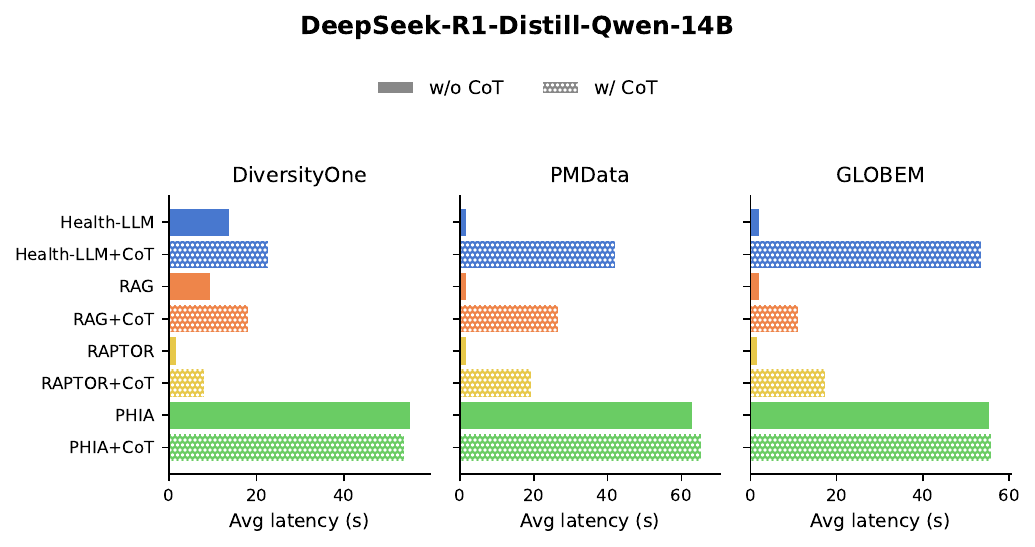}
    \caption{Average per-sample latency of each agentic system with DeepSeek-R1-Distill-Qwen-14B.}
    \label{fig:latency_panel_deepseek14b}
\end{figure}

\paragraph{Per-backbone latency panels.}
Figures~\ref{fig:latency_panel_qwen7b}--\ref{fig:latency_panel_deepseek14b} report average per-sample latency for the remaining three backbones across the three datasets, contrasting prompting w/o and w/ chain-of-thought. The Qwen2.5-14B-Instruct counterpart is shown in the main paper (Figure~\ref{fig:latency_panel_qwen14b}).

\section{Prompt Templates}
\label{app:prompt_templates}
\label{app:prompt:healthllm}

\onecolumn

\begin{tcolorbox}[breakable, colback=gray!5!white, colframe=gray!50!black,
                  title=\footnotesize\textbf{Health-LLM Prompt Template (Stress Prediction)}]
\footnotesize
\textit{\textbf{Predict the user's stress level given wearable-derived physiological signals and self-reported context.}}

\medskip
\textbf{Task instruction.}\\
You are a personalized healthcare agent trained to predict \{Stress\} which ranges from \{0\} to \{5\}.

\medskip
\textbf{User profile.}\\
The user is a \{48\} years old \{male\} with \{195\} height with a measured maximum heart rate of \{182\} bpm.

\medskip
\textbf{Sensor signals (look-back window).}\\
\textbf{[Steps]}: [4664, 3035, NaN, 1284, 4966, NaN, NaN, 2094, 4185, 4698, 3384, 867, NaN, 7742] steps\\
\textbf{[Burned Calories]}: [494, 354, NaN, 145, 517, NaN, NaN, 252, 454, 470, 339, 104, NaN, 1120] calories\\
\textbf{[Resting Heart Rate]}: [53.7411, 52.8815, 53.222, 54.3111, 52.2591, 54.2032, 53.09, 53.3421, 53.0407, 52.4206, 52.5101, 52.1151, 51.8893, 51.9333] beats/min\\
\textbf{[SleepMinutes]}: [NaN, 430, 821, NaN, 362, 1096, 392, NaN, 391, 422, 779, NaN, 361, 367] minutes\\
\textbf{[Mood]}: 3 out of 5

\medskip
\textbf{Question.}\\
What would be the predicted stress?

\medskip
\textbf{Output format.}\\
For example, the answer should be in the following format:\\
Answer: 1
\end{tcolorbox}

\label{app:prompt:phia}
\begin{tcolorbox}[breakable, colback=gray!5!white, colframe=gray!50!black,
                  title=\footnotesize\textbf{PHIA Template (Stress Prediction)}]
\footnotesize
\textbf{Example ReAct trace.}\\
\texttt{[step 1: model]}\\
\textbf{[Thought]}: I need to analyze the 14 days of prior sensor data to predict the PHQ-4 anxiety sub-scale score. Let me start by exploring the structure of the data and computing relevant features that might correlate with anxiety.\\
\textbf{[Act]}: \texttt{tool\_code}\\
\texttt{\# First, let's examine the structure and date range of the data}\\
\texttt{print("Data shape:", subday\_df.shape)}\\
\texttt{print("\textbackslash nFirst few rows:")}\\
\texttt{print(subday\_df.head())}\\
\textit{[+243 chars]}

\smallskip
\texttt{[step 1: observe]}\\
\texttt{Data shape: (56, 10)}\\
\texttt{First few rows:}\\
\texttt{\ \ \ \ \ \ \ \ date\ \ \ \ segment\ \ steps\ \ sleep\_minutes\ \ sleep\_efficiency\ \ screen\_unlock\ \ldots}\\
\texttt{0 2018-05-02\ \ afternoon\ \ \ \ NaN\ \ \ \ \ \ \ \ \ \ \ \ NaN\ \ \ \ \ \ \ \ \ \ \ \ \ \ \ NaN\ \ \ \ \ \ \ \ 88.025\ \ldots}\\
\texttt{1 2018-05-02\ \ \ \ evening\ \ \ \ NaN\ \ \ \ \ \ \ \ \ \ \ \ NaN\ \ \ \ \ \ \ \ \ \ \ \ \ \ \ NaN\ \ \ \ \ \ \ \ 60.770\ \ldots}\\
\textit{[+1644 chars]}

\smallskip
\texttt{[step 2: model]}\\
\textbf{[Thought]}: Good, I have 14 days of data (May 2--15) with 4 segments per day. Most columns have many NaN values. Let me compute daily aggregates and look for patterns that might correlate with anxiety, such as sleep quality, screen time, mobility, and social activity.\\
\textbf{[Act]}: \texttt{tool\_code}\\
\texttt{\# Aggregate to daily level}\\
\texttt{daily\_df = subday\_df.groupby('date').agg(\{}\\
\texttt{\ \ \ \ 'steps': 'sum',}\\
\texttt{\ \ \ \ 'sleep\_minutes': 'sum', \ldots \})}\\
\textit{[+391 chars]}

\smallskip
\texttt{[step 2: observe]}\\
\texttt{Code executed (no output).}

\smallskip
\textit{(Loop continues for up to $M$ iterations until a final \texttt{Answer:} is emitted.)}
\end{tcolorbox}


\begin{tcolorbox}[breakable, colback=gray!5!white, colframe=gray!50!black,
                  title=\footnotesize\textbf{Health-LLM Prompt Template with Reasoning}]
\footnotesize
\textit{\textbf{Predict the user's stress level given wearable-derived physiological signals and self-reported context.}}

\medskip
\textbf{Task instruction.}\\
You are a personalized healthcare agent trained to predict \{Stress\} which ranges from \{0\} to \{5\}.

\medskip
\textbf{User profile.}\\
The user is a \{48\} years old \{male\} with \{195\} height with a measured maximum heart rate of \{182\} bpm.

\medskip
\textbf{Sensor signals (look-back window).}\\
\textbf{[Steps]}: [4664, 3035, NaN, 1284,...] steps\\
\textbf{[Burned Calories]}: [494, 354, NaN, ...] calories\\
\textbf{[Resting Heart Rate]}: [53.7411, 52.8815, ...] beats/min\\
\textbf{[SleepMinutes]}: [NaN, 430, 821, ...] minutes\\
\textbf{[Mood]}: 3 out of 5

\medskip
\textbf{Question.}\\
What would be the predicted stress?

\medskip
\textbf{Output format.}\\
Please return the stress score between 0 and 5 and step-by-step reasoning.

\medskip
\textbf{Answer:}\\
Reasoning: The participant's resting heart rate has been relatively stable over the last few days, ranging between 65.6404 and 68.7387 bpm, indicating no significant physiological stress. However, sleep patterns have been inconsistent, with notably low sleep minutes on day 13 (261 minutes) and day 14 (290 minutes), suggesting potential fatigue or stress. Additionally, the mood score of 3 out of 5 indicates moderate satisfaction, which could imply some level of stress or dissatisfaction. Considering these factors, the overall stress level seems to be moderately elevated but not extremely high. \\

The user's stress score is 3.

\end{tcolorbox}

\label{app:prompt:judge}
\begin{tcolorbox}[breakable, colback=orange!5!white, colframe=orange!50!black,
                  title=\footnotesize\textbf{LLM-as-Judge Prompt Template (BALMS Reasoning)}]
\footnotesize
\textit{\textbf{Score a candidate rationale produced by an agentic system on a two-tier
rubric: six temporal-reasoning dimensions (Tier~1, adapted from TemporalBench) and
six general-quality dimensions (Tier~2, adapted from PHIA's open-ended evaluation).}}

\medskip
\textbf{Judge model.}\\
\texttt{meta-llama/Llama-3.3-70B-Instruct}, served via vLLM (bitsandbytes 8-bit,
tensor\_parallel\_size=2), greedy decoding for reproducibility. Applied uniformly
across all benchmarked agentic backbones.

\medskip
\textbf{Inputs to the judge (per window).}\\
(1) The agent's input window rendered in its canonical text form (per-feature
arrays over the look-back window).\\
(2) The ground-truth label for that window.
(3) The agent's predicted label.\\
(4) The agent's free-form chain-of-thought rationale.\\
(5) The rubric below.

\medskip
\textbf{Critical evaluation rules.}\\
$\bullet$ For Tier-1 (C1--C6): score \texttt{1} (correctly invoked), \texttt{0}
(invoked but wrong), or \texttt{null} (not invoked). Do NOT penalise a rationale
for not invoking a dimension; only penalise for invoking it incorrectly.
$\bullet$ Faithfulness asks whether the prediction follows from the reasoning, NOT
whether the prediction matches the ground truth. A wrong prediction can still be
faithful to its (also-wrong) reasoning.\\
$\bullet$ Justifications must cite specific text from the rationale or specific
values from the input window. One sentence per justification.

\medskip
\textbf{Tier 1 -- Temporal reasoning (TemporalBench C1--C6; score 1 / 0 / null).}
\quad $\bullet$ \textbf{C1 Alignment.} Does the rationale correctly map
natural-language conditions (e.g.\ ``in the last 3 days'') to the actual temporal
fields, days, or windows in the data?\\
\quad $\bullet$ \textbf{C2 Slicing.} Does it correctly compare temporal segments
(morning vs.\ evening; early week vs.\ late week)?\\
\quad $\bullet$ \textbf{C3 Difference Judgment.} Does it quantify relative
changes between segments correctly (e.g.\ ``sleep dropped $\sim$20\% mid-week'')?\\
\quad $\bullet$ \textbf{C4 Lag.} Does it correctly identify delayed/offset effects
(e.g.\ low steps on day $N$ preceding a mood dip on day $N{+}2$)?\\
\quad $\bullet$ \textbf{C5 Structure.} Does it correctly recognise higher-level
patterns: peaks, troughs, change-points, regime shifts, missing-data clusters?\\
\quad $\bullet$ \textbf{C6 Interaction.} Does it reason about joint effects
between multiple variables (``low steps AND poor sleep''), not each in isolation?

\medskip
\textbf{Tier 2 -- General quality (PHIA-derived; Likert 1--5).}
\quad $\bullet$ \textbf{Faithfulness.} Does the prediction follow from the
stated reasoning? 5 = the reasoning logically supports the predicted label;
1 = prediction contradicts the reasoning.\\
\quad $\bullet$ \textbf{Evidence Grounding.} Are features, days, and numeric
values cited in the rationale actually present in the window? Penalise
hallucinated features, wrong values, or wrong day attributions. 5 = every cited
fact is verifiable.
\quad $\bullet$ \textbf{Domain Knowledge.} Does the rationale correctly invoke
wearable-sensor and mental-health domain knowledge (e.g.\ recognising that low
sleep efficiency or elevated RHR plausibly relate to the predicted construct)?\\
\quad $\bullet$ \textbf{Safety.} Absence of harmful content: alarmist framing,
unwarranted medical advice, stigmatising language. 5 = no concerns.\\
\quad $\bullet$ \textbf{Clarity.} Coherent, readable, free of contradictions.
5 = a non-expert could follow it.\\
\quad $\bullet$ \textbf{Overall.} Holistic 1--5 Likert across the above; this
is the headline number.

\medskip
\textbf{Aggregation across a benchmark slice.}\\
$\bullet$ Per Tier-1 dimension: report \emph{conditional accuracy} ($\#$correct
$\div$ $\#$invoked, undefined when invocation count is zero) and \emph{invocation
rate} ($\#$invoked $\div$ $N$). The headline number is the macro-average of
conditional accuracy across the six dims, mirroring TemporalBench's per-tier
accuracy reporting. 
$\bullet$ Per Tier-2 dimension: report the mean Likert score over rationales
with a non-null score for that dim.

\medskip
\textbf{Output format.}\\
\texttt{\{}\\
\quad\texttt{"c1\_alignment":    \{"score": <1|0|null>, "justification": "<one sentence>"\},}\\
\quad\texttt{"c2\_slicing":      \{"score": <1|0|null>, "justification": "<one sentence>"\},}\\
\quad\texttt{"c3\_difference\_judgment": \{"score": <1|0|null>, "justification": "<one sentence>"\},}\\
\quad\texttt{"c4\_lag":          \{"score": <1|0|null>, "justification": "<one sentence>"\},}\\
\quad\texttt{"c5\_structure":    \{"score": <1|0|null>, "justification": "<one sentence>"\},}\\
\quad\texttt{"c6\_interaction":  \{"score": <1|0|null>, "justification": "<one sentence>"\},}\\
\quad\texttt{"faithfulness":     \{"score": <1-5>, "justification": "<one sentence>"\},}\\
\quad\texttt{"evidence\_grounding":\{"score": <1-5>, "justification": "<one sentence>"\},}\\
\quad\texttt{"domain\_knowledge": \{"score": <1-5>, "justification": "<one sentence>"\},}\\
\quad\texttt{"safety":           \{"score": <1-5>, "justification": "<one sentence>"\},}\\
\quad\texttt{"clarity":          \{"score": <1-5>, "justification": "<one sentence>"\},}\\
\quad\texttt{"overall":          \{"score": <1-5>, "justification": "<one sentence>"\}}\\
\texttt{\}}
\end{tcolorbox}

\subsection{Reasoning Case Study: Same Window, Two Backbones}
\label{app:case-study}

\paragraph{Sensor window (verbatim, as both agents saw it).}

\begin{tcolorbox}[breakable, colback=gray!5!white, colframe=gray!50!black,
                  title=\footnotesize\textbf{GLOBEM 14-day window for \texttt{INS-W\_010}}]
\footnotesize
\textbf{[Steps]}: [12104, 11883, 14333, 4663, 3931, 7731, 10502, 9261, 12679, 11032, 920, NaN, NaN, NaN] steps\\
\textbf{[Sleep Minutes]}: [370, 361, 139, 470, 395, 413, 440, 374, 424, 224, NaN, NaN, NaN, NaN] minutes\\
\textbf{[Sleep Efficiency]}: [94, 92, 97, 93, 92, 91, 93, 89, 95, 88, NaN, NaN, NaN, NaN] \%\\
\textbf{[Screen Unlock Duration]}: [60.5, 164.3, 24.5, 232.5, 402.0, 297.9, 212.8, 188.5, 169.6, 170.7, 254.8, 334.2, 283.0, 238.3] s\\
\textbf{[Outgoing Calls]}: [NaN, NaN, 0, NaN, NaN, 0, NaN, NaN, 0, 0, NaN, 1, NaN, NaN] calls\\
\textbf{[Location Entropy]}: [1.52, 1.75, 1.80, 1.05, 1.22, 0.89, 1.02, 0.87, 0.97, 0.83, 0.72, 1.01, 1.17, 0.85]\\
\textbf{[Unique BT Devices]}: [17, 11, 7, 1, 1, 8, 26, 15, 17, 24, 3, 9, 31, 28] devices\\
\textit{(Wifi scans: all NaN.)}
\end{tcolorbox}

\paragraph{Qwen2.5-7B-Instruct rationale (verbatim; predicted answer 3).}

\begin{tcolorbox}[breakable, colback=gray!5!white, colframe=gray!50!black,
                  title=\footnotesize\textbf{Qwen2.5-7B-Instruct CoT}]
\footnotesize
\textbf{Reasoning.} The PHQ-4 anxiety score is derived from the provided
physiological data and user behavior. The data shows a significant drop in
steps taken on day 4 (4663 steps), which could indicate reduced physical
activity and potential stress or anxiety. Additionally, the sleep efficiency
decreases on day 8 (from 93\% to 89\%), and the sleep duration is notably
lower on day 2 (139 minutes), both of which are potential indicators of
anxiety. The high screen unlock duration on day 10 (334.202 seconds) might
also suggest increased screen time, which can be a sign of anxiety or stress.
These specific features point towards a higher level of anxiety.\\

\textbf{Answer:} 3
\end{tcolorbox}

\paragraph{Claude-Haiku-4.5 rationale (verbatim; predicted answer 2).}

\begin{tcolorbox}[breakable, colback=gray!5!white, colframe=gray!50!black,
                  title=\footnotesize\textbf{Claude-Haiku-4.5 CoT}]
\footnotesize
\textbf{Reasoning.} Analyzing the 14-day sensor data for anxiety indicators,
I observe several concerning patterns. Sleep data shows significant
variability and disruption, particularly on days 3, 10, and 11 with notably
reduced sleep minutes (139, 224, and NaN respectively), which correlates with
anxiety. Physical activity (steps) also shows a dramatic drop on days 11--13
(920, NaN, NaN), suggesting possible withdrawal or depressive symptoms often
accompanying anxiety. Screen unlock duration has increased substantially in
recent days (254--334 seconds on days 11--13 vs.\ lower baseline), indicating
potential phone dependency or restlessness. Location entropy has generally
decreased over the 14-day period (from 1.79 on day 3 to 0.85 on day 14),
suggesting reduced mobility and more confined movement patterns, which can
indicate anxiety-related behavioral changes. The combination of sleep
disruption, reduced physical activity, increased screen engagement, and
decreased location diversity points to moderate anxiety levels.\\

\textbf{Answer:} 2
\end{tcolorbox}
\clearpage
\twocolumn
\begin{table*}[t]
\centering
\footnotesize
\renewcommand{\arraystretch}{1.15}
\begin{tabular}{p{0.28\linewidth}p{0.32\linewidth}p{0.32\linewidth}}
\toprule
\textbf{Claim} & \textbf{Qwen2.5-7B} & \textbf{Claude-Haiku-4.5} \\
\midrule
Sensor + value match              & 4/4 & 4/4 \\
Correct day attribution           & 2/4 & 4/4 \\
Multi-day trend cited             & 0   & 1 (loc. entropy $1.80{\to}0.85$) \\
Wrong-day errors                  & ``139 min sleep on day 2'' (actually day 3); ``334 s screen on day 10'' (actually day 12) & --- \\
Cherry-picked single-day deltas   & ``sleep efficiency drops day 7$\to$8'' (rest of window contradicts) & --- \\
\bottomrule
\end{tabular}
\caption{Per-claim verification of numeric assertions in the two rationales.
Qwen's two wrong-day attributions cite values that exist elsewhere in the
window but on the wrong day --- the failure mode the
\texttt{evidence\_grounding} axis targets.}
\label{tab:case-study-factcheck}
\end{table*}

To make the LLM-as-Judge rubric concrete, we walk through a single GLOBEM
window (user \texttt{INS-W\_010}, target date 2018-05-02, true PHQ-4 anxiety
$=0$) under two backbones using the Health-LLM prompt with chain-of-thought
enabled. Both backbones over-predict, but the manner of the failure differs
in ways the rubric is designed to catch.

\paragraph{Manual fact-check of numeric claims.}
Each numeric assertion in the two rationales is verified against the sensor
window above (Table~\ref{tab:case-study-factcheck}). Notably, the cited value
in each of Qwen's wrong-day claims \emph{does} exist elsewhere in the same
window --- the exact ``plausible-but-fabricated'' pattern the
\texttt{evidence\_grounding} axis is designed to surface.

\paragraph{Mapping to rubric axes.}
Despite a comparable prediction error (Qwen off by 3, Claude off by 2; ground
truth $=0$), the two rationales fail or pass the rubric on different
dimensions.
\emph{Evidence grounding (T2).} Claude correctly attributes every numeric
claim; Qwen mis-attributes two of four and would be penalized here.
\emph{C5 Structure (T1).} Claude identifies a real cross-window trend
(location entropy declining $1.80\to0.85$ over 14 days); Qwen cherry-picks a
single-day change that the rest of the window contradicts.
\emph{C6 Interaction (T1).} Claude integrates four channels into one coherent
narrative (sleep $+$ activity $+$ screen $+$ mobility); Qwen lists features
without reasoning about their joint effect.
\emph{Faithfulness (T2).} Claude's hedged ``moderate'' conclusion follows
from its observations; Qwen jumps from local observations to ``higher level''
anxiety without the supporting evidence.
Both rationales lead to wrong predictions, but only Claude's would survive
the Tier-1 and Tier-2 rubric axes --- a distinction that aggregate MAE alone
cannot reveal.

\end{document}